# An AI model for Rapid and Accurate Identification of Chemical Agents in Mass Casualty Incidents

Nicholas Boltin[1], Daniel Vu[1], Bethany Janos[1], Alyssa Shofner[1], Joan Culley[2] and Homayoun Valafar[1*]

[1]Department of Computer Science and Engineering, University of South Carolina, Columbia, SC

[2]College of Nursing, University of South Carolina, Columbia, SC

*Corresponding Author: homayoun@cse.sc.edu

**Abstract** – *In this report we examine the effectiveness of WISER in identification of a chemical culprit during a chemical based Mass Casualty Incident (MCI). We also evaluate and compare Binary Decision Tree (BDT) and Artificial Neural Networks (ANN) using the same experimental conditions as WISER. The reverse engineered set of Signs/Symptoms from the WISER application was used as the training set and 31,100 simulated patient records were used as the testing set. Three sets of simulated patient records were generated by 5%, 10% and 15% perturbation of the Signs/Symptoms of each chemical record. While all three methods achieved a 100% training accuracy, WISER, BDT and ANN produced performances in the range of: 1.8%-0%, 65%-26%, 67%-21% respectively. A preliminary investigation of dimensional reduction using ANN illustrated a dimensional collapse from 79 variables to 40 with little loss of classification performance.*

## 1 Introduction

Improvement of the healthcare system in the United States is the subject of great interest and debate in the social, political, and economical arenas of our society. One obvious approach in improving the overall healthcare system is by eliminating the existing inefficiencies that impede our system[1–3]. Removal of inefficiencies impacts our system of healthcare in two inherent ways: significant improvement of the patient outcome, and a reduction in the cost of healthcare. Although in principle it is clear that removal of inefficiencies is beneficial, in practice there has been little effort in removal of the existing inefficiencies. This lack of effort is rooted in the complexity of our healthcare system that has manifested itself as a lack of consensus on the method of removing the existing inefficiencies.

Integration of technological advances in our healthcare such as utilization of mobile devices, availability of broadband systems with high throughput, and embedded clinical decision systems[4–6] can be cited as some approaches that can reduce overall inefficiencies of our healthcare system. One branch of healthcare that can benefit from better streamlining of patient-care through integration of clinical decision support is in emergency care during a mass casualty incident[7] (MCI). The rapid operational tempo of an Emergency Room (ER) serves as an ideal vehicle to study any existing inefficiencies while the resource-limited conditions of an MCI will help in clearly gauging the impact of any proposed improvements. MCI events clearly require rapid treatment of patients with minimum interruption for data collection, while optimal treatment of patients requires the hindering and cumbersome completion of detailed patient information to identify the culprit chemical substance. These two competing objectives have traditionally been a major impediment in optimizing the MCI treatment process with a natural priority extended to rapid treatment of patients. Therefore, there has been little advances in improving treatment of chemical MCI events. Research is needed to build a better understanding of the information and technological needs of the healthcare and public health workforce during emergency decision making[8].

A limited set of clinical decision support software have been introduced by the larger community[9]. The National Library of Medicine has created the Wireless Information System for Emergency Responders[10] (WISER), which allows emergency responders to identify a list of possible chemical substances based on the observed patient symptoms. The US Department of Health and Human Services has developed another software tool named the Chemical Hazards Emergency Medical Management-Intelligent Syndromes Tool[11] (CHEMM-IST). CHEMM-IST is a prototype that guides first-responders through a series of questions related to signs and symptoms that leads to a probabilistic diagnosis of four syndromes rather than a list of chemical hazards. Although such software make significant strides in assisting the process of emergency care, their efficacy have not been assessed during a chemical based MCI.

In this report we examine the effectiveness of WISER as the potential software for early identification of chemical material during an MCI event using simulated patient signs/symptoms (SSx) that we have reverse engineered from WISER. We also report results from Binary Decision Tree and Artificial Neural Network applications to the same set of simulated patient data. We conclude by reporting results of our initial investigation aimed at dimensional reduction of SSx space. Our final objective is to challenge the paradigm that rapid patient treatment is in contrary to data gathering that will assist in early identification of culprit chemical. We contest that careful design of sophisticated clinical decision support tools can satisfy both competing objectives of rapid information gathering and accurate chemical identification processes.

## 2 Materials and Methods

Our general approach consists of creating signs and symptoms (SSx) for simulated patients using a reverse-engineered table of SSx from the WISER application. Using the simulated data, we then proceed to evaluate the successful identification of a culprit chemical using WISER, Binary Decision Tree (BDT), and Artificial Neural Network (ANN) machine learning approaches.

### 2.1 WISER

Wireless Information System for Emergency Responders[10] (WISER) is a free application available for Android and iOS, which can also be downloaded as a standalone application on a desktop computer. Developed by the National Library of Medicine (NLM), WISER is a system designed to assist emergency responders in hazardous material incidents. It provides a wide range of information on hazardous substances, including substance identification support, physical characteristics, human health information and containment and suppression advice. Its key features include rapid access to the most important information about a hazardous substance by an intelligent synopsis engine and display called "Key Info", and access to NLM's Hazardous Substances Data Bank (HSDB), which contains detailed peer-reviewed information on hazardous substances and comprehensive decision support.

The key feature in WISER most relevant to this work is the Substance ID Support (SIDS). This allows an emergency responder to input patient SSx, from which the SIDS will identify one or more likely hazardous chemicals causing those symptoms. WISER contains a checklist of 79 SSx, which are input for selected systems of the body through an interactive tool as seen in Figure 1A. As the signs and symptoms are entered (Figure 1B) the pre-populated library of 438 hazardous substances is successively reduced. The user can view the list, select a substance and view toxicology information available in the HSDB, which contains data from the NLM Toxicology Data Network[12] (TOXNET). The HSDB data file contains information on human exposure, industrial hygiene, emergency handling procedures, environmental fate, regulatory requirements and related area.

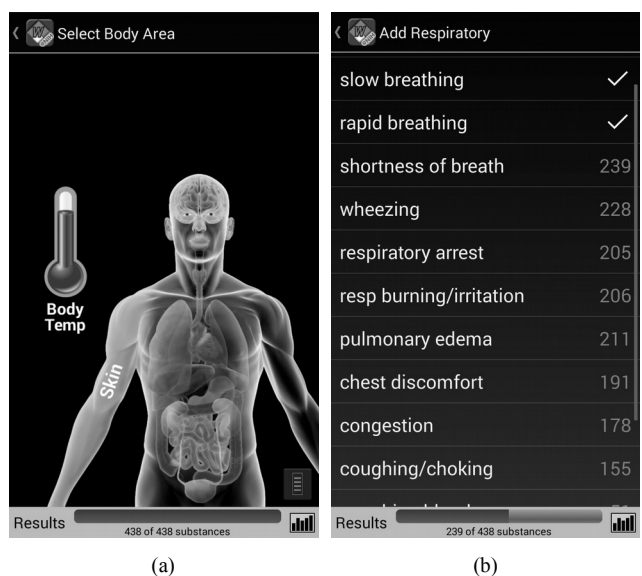

Figure 1. Wireless Information System for Emergency Responders (WISER) for Android operating system. Panel (a) is the Interactive tool and panel (b) is the symptom selection interface. Panel (b) also shows the substance ID support in which an emergency responder can identify an unknown substance based on signs and symptoms of victims.

### 2.2 Reverse engineering and compression of WISER database

A thorough evaluation of WISER necessitated reverse engineering of all WISER's substances with their associated SSx. This task was performed by manually reviewing NLM's HSDB and parsing the SSx for each substance. An example of the resultant table of SSx is shown in Figure 2. Each of the 438 substances found in WISER is represented in the first column in this table, and the following 79 columns represent the corresponding SSx found in WISER for a given chemical. The presence or absence of each SSx is indicated by a 1 or a 0 respectively.

Figure 2. WISER's reconstructed database using NLM's toxicology information stored in the Hazardous Substances Data Bank (HSDB).

Examination of the created database revealed several substances with identical SSx profiles. In such instances, a cluster of chemicals was reduced to a single representative. The list of uniquely distinguishable chemicals was then reduced from 438 substances to 311 unique substances, which serves as the reverse-engineered list of unique chemicals.

## 2.3 Creation of Simulated Victims (Test Set)

Simulated patient-data were generated from the ideal database of 311 unique substances by perturbation of randomly selected Ssx. This was done to precisely control the amount of missing data. Signs and symptoms related to a real MCI would be ideal, however accurate patient records during these scenarios are limited and usually incomplete[13,14]. Each substance was replicated 100 times to create a reasonably extensive testing set that consisted of 31,100 simulated victims. Three data-sets were created by random toggling of selected SSx at 5%, 10%, and 15% selection rates. To ensure the proper random selections, probability density profiles were examined for the number of perturbed SSx across each of the simulated patient-data. An overview of the perturbed data-sets (shown in Figure 3) corroborates the intended rates of perturbation.

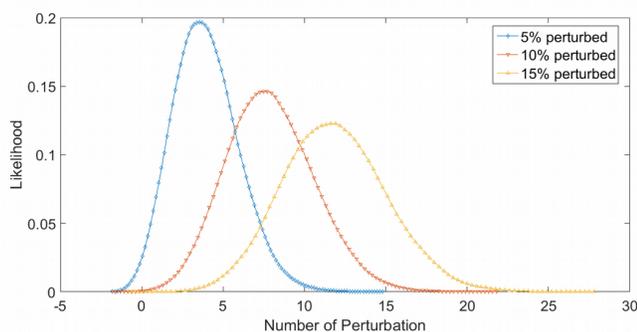

Figure 3. The Kernel Density Estimation of the 3 test data-sets. Test data were created by starting with the ideal table of symptoms from WISER and changing the symptoms by 5%, 10%, and 15%.

## 2.4 Overview of Machine Learning Approach

Our general work-flow for creating predictive models can be found in Figure 4. Supervised machine learning techniques were utilized in the Matlab 2015Rb environment to identify patterns and to develop predictive models. Our process began by importing the reverse-engineered database of 311 unique substances followed by training of two types of classification models: Binary Decision Trees (BDT) and Artificial Neural Networks (ANN). After successful training of a given model, the known SSx profile for all 311 substances was tested on the trained model to establish proper learning (testing for memorization versus generalization is conducted in a different step). The model with the highest accuracy during the training was chosen as the final model. Evaluation of each trained model was then assessed using the SSx profiles of the 31,100 simulated victims. A prediction accuracy was calculated as shown in Equation 2. In this equation $A$ represents the accuracy of the model (expressed in %), $N_c$ indicates the number of correctly identified chemicals, and $N_{total}$ represents the total number of trials (31,100 in this case). The next sections provide a more detailed description of the training and testing for each model.

$$A = (N_c / N_{Total}) \cdot 100 \qquad (1)$$

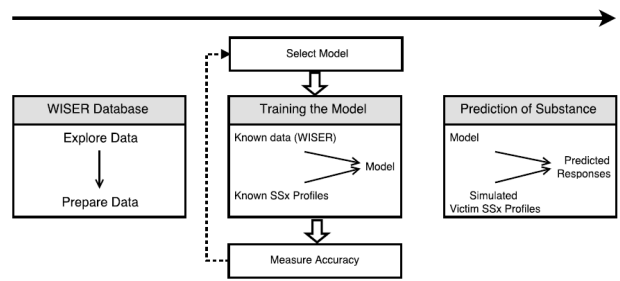

Figure 4. Work-flow for exploration of data, training models and predicting substances using supervised machine learning techniques

## 2.5 Training and Testing of Classification Methods

We evaluated three common classification approaches in our investigation. The classification approaches consisted of: database look-up (as implemented by WISER), Binary Decision Trees, and Artificial Neural Networks. Details for each of the three approaches are described in the following sections.

### 2.5.1 Database look-up (WISER)

The interactive nature of WISER was the limiting factor in automated and batch evaluation of WISER for 31,100 patients each represented by 79 SSx. This limitation served as one of our primary motivations in establishing a local database of WISER SSx. The first step in replicating a process identical to the WISER application was to understand its selection logic. WISER selects chemicals only based on the presence of a SSx and not its absence. Therefore, WISER will identify the entire library of 438 (or 311 unique) chemicals as the potential list of possible exposed chemicals for a patient exhibiting no apparent SSx. While this logic may appear questionable in our application, we proceeded with our evaluation of WISER in an exact fashion. Our initial evaluation of WISER consisted of a query-based search of our local database of chemicals using MySQL database engine housed on an Ubuntu LTS 14.04 server. This approach required a database look-up for SSx of all 31,100 simulated patients. Since the WISER approach may (and most likely will) return a list of potential chemicals, the database look-up step is followed by a search for existence of the right chemical in the list of returned chemicals. Although the time requirement of this evaluation mechanism was feasible (in the order of a week) for a list of 31,100 patients, it is an impractical approach for future investigations with larger data-sets in order to establish a more thorough evaluation of the methods. Our most current approach consists of an in-house developed program to simulate this table look-up process. Our evolved approach returns the identical results that WISER would return while reducing the search time from months to seconds. Our testing process consisted of recording the number of times that the correct chemical was present in the list of returned chemicals similar to Equation 1.

Since WISER operates in a deterministic fashion, a statistical model of its performance can be developed. By assuming that every patient will undergo alteration of exactly $n$ SSx, it can be argued that WISER's outcome should closely follow a success rate shown in Equation 2. This equation lists all of the possible perturbation of SSx that will result in removal of the correct chemical in WISER's resultant list. This equation can be simplified using the Binomial theorem as shown in Equation 2. Based on binomial distribution modeling of the WISER's outcome, a success rate of 6.25%, 0.4% and 0.02% can be expected for the cases of 5%, 10% and 15% perturbation of SSx.

$$r = 1 - \sum_{i=1}^{n} \binom{n}{i} p^i (1-p)^{(n-i)} = \frac{1}{2^n} \qquad (2)$$

### 2.5.2 Binary Decision Tree

A Binary Decision Tree (BDT) was trained using the reverse-engineered WISER database within the Matlab 2015Rb environment. A maximum deviance reduction was used as the split criterion with 350 maximum splits. Each of the 311 chemicals was replicated 312 times to facilitate the construction of a complete tree and in consideration of Matlab's training algorithm. Under this training conditions, a classification rate of 100% was achieved.

Our adopted testing procedure consisted of observing the chemical identification accuracy of the trained network with the simulated patient-data. It is noteworthy that the trained BDT was based on ideal data while the testing was based on the perturbed data-sets (5%, 10% and 15% perturbation).

### 2.5.3 Artificial Neural Network

An Artificial Neural Network (ANN) was trained through the Pattern Recognition toolbox of the Matlab 2015Rb using back-propagation learning algorithm[15–17]. The unique set of 311 ideal chemical SSx were used during the training of the ANNs. The training set consisted of 5 identical replicas for each of the unique 311 chemicals (for a total of 1555 training patterns) in order to accommodate a random selection of the cross-validation and testing sets. The 1555 training patterns were randomly partitioned into 70% for training, 15% for cross-validation and 15% for testing. Numerous ANNs were trained and tested for selection of the optimal number of hidden neurons. Our investigation concluded 20 neurons as the optimal number of hidden neurons. The final trained ANN model exhibited cross-entropy results of: 4.4 for the training set, 12.7 for the cross-validation set, and 12.7 for the testing set. These outcomes correspond to: 0% error for the training set, 2.1% error for the validation set and 2.1% for the testing set.

To test the performance of the network with unknown data, the 31,100 simulated patient-data were used as inputs for the ANN trained with ideal chemical data.

## 3 Results and Discussion

### 3.1 Database look-up (WISER)

The results of WISER database look-up approach are shown in Table 1 and exhibit a reasonable correlation to the binary distribution model shown in Equation 2. The rapid decay in performance of WISER is easily expected. We use the results of WISER as the basis of comparison since it is the most prominent and existing mechanism.

*Table 1: Prediction accuracy results from WISER testing using 31,100 simulated patient-data perturbed at 5%, 10% and 15%.*

| Data-set | Prediction Accuracy | Max | Min |
|---|---|---|---|
| 5% Perturbed | 1.8% | 7% | 0% |
| 10% Perturbed | 2.3x10$^{-2}$% | 1% | 0% |
| 15% Perturbed | 0.0% | 0% | 0% |

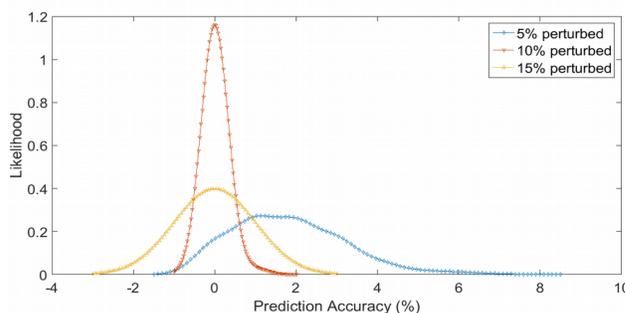

Figure 5: The Kernel Density Estimations from testing WISER with 31,100 simulated patient-data perturbed at 5%, 10% and 15%

### 3.2 Binary Decision Tree

Testing results for BDT are shown in Table 2. In this table the first columns represent the severity of the perturbation and the second column corresponds to the classification accuracy of the BDT. The third and fourth columns of Table 2 list the minimum and maximum performance across all of the 311 chemical substances. To better understand the performance of the BDT across the entire ensemble of 311 chemicals, a probability density function was created using the Kernel Density Estimation (KDE) technique[15,18]. Figure 6 illustrates the statistics for BDT classification behavior over the entire 100 representatives of each 311 chemicals. The nearly Gaussian distribution of the statics indicate a very well behaved system without any particular bias.

Another important factor to monitor during the construction of a BDT is the topology of the final tree. Figure 7 illustrates the topology of the final tree (in the interest of simplicity the labels are omitted), which indicates a very well balanced tree of depth 9. This depth is in perfect theoretical agreement with the complexity of the problem, serving as another indication of a successful training session.

*Table 2: Prediction accuracy results for Binary Decision Tree (BDT) testing using 31,100 simulated patient-data perturbed at 5%, 10% and 15%.*

| Data-set | Prediction Accuracy | Max | Min |
|---|---|---|---|
| 5% Perturbed | 64.9% | 81% | 53% |
| 10% Perturbed | 41.8% | 54% | 27% |
| 15% Perturbed | 25.6% | 40% | 13% |

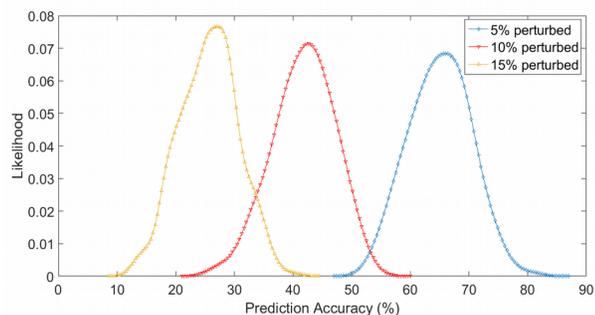

Figure 6.The Kernel Density Estimations from testing the Binary Decision Tree (BDT) model with 31,100 simulated patient-data perturbed at 5%, 10% and 15%.

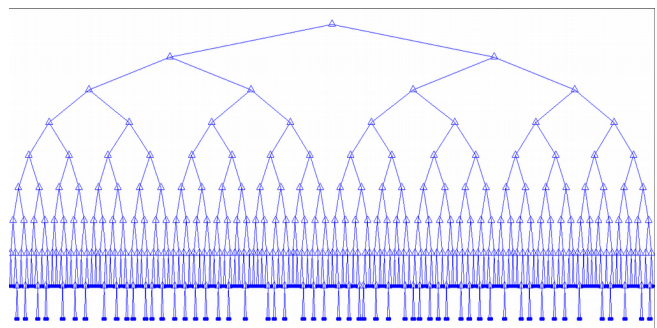

Figure 7.Static binary decision tree for 311 unique chemicals found in the National Library of Medicine's Hazardous Substances Data Bank (HSDB).

### 3.3 Artificial Neural Networks

The evaluation results of the ANN are shown in Table 3. Similar to the results of BDT, the first two columns of this table indicate the severity of perturbation and outcome accuracy, while columns three and four indicate the range of the outcomes across all 311 chemicals. Remarkably the accuracy of BDT and ANN appear to be similar, while the range of ANN's performance exhibit a larger variation. To better understand the statistics of ANN's results, probability density profiles were created for each of the experiment using KDE using the exact parameters as the BDT (identical kernels). Similar to BDT, the Gaussian nature of the outcomes indicate a well behaved and unbiased system. Visual inspection of Figure 8 confirms the noted differences in variation of outcomes compared to the BDT results.

*Table 3: Prediction accuracy results for the Artificial Neural Network (ANN) testing using 31,100 simulated patient-data perturbed at 5%, 10% and 15%.*

| Data-set | Prediction Accuracy | Max | Min |
|---|---|---|---|
| 5% Perturbed | 67.2% | 96% | 28% |
| 10% Perturbed | 38.4% | 73% | 10% |
| 15% Perturbed | 21.4% | 49% | 3% |

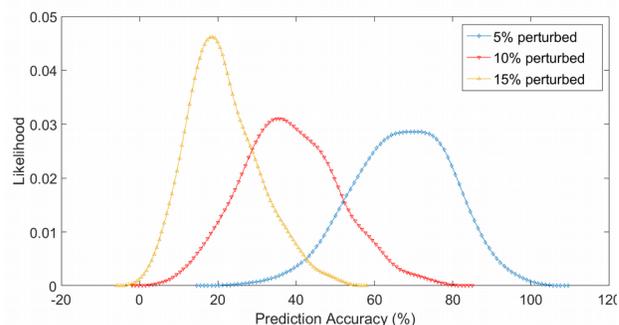

Figure 8.The Kernel Density Estimations from testing the Artificial Neural Network (ANN) model with 31,100 simulated patient-data perturbed at 5%, 10% and 15%.

### 3.4 Dimensional reduction

To optimize the Artificial Neural Network model, we examined the number of hidden neurons being used during the training phase of the model development. 10 models were trained, each with a different number of hidden neurons starting with 10 hidden neurons, then incrementing by 10 and the final model using 100 hidden neurons. After the model was created, additional testing was performed using the 5% perturbed data-set and the amount of error from the ANN was recorded. As seen in Figure 9, the results show that as we increase the number of hidden neurons, the amount of error from the ANN is reduced with the minimal amount of error being 15.4% at 100 hidden neurons. We then examined training the ANN with only the first 40 SSx instead of the complete database of 79 SSx. Again 10 models were trained starting with 10 hidden neurons at increments of 10 to 100 hidden neurons. After training the ANN, additional testing was also performed using the 5% perturbed data-set and recording the ANN error. It can be seen in Figure 9, the results followed the same pattern as with 79 SSx with the minimal amount of error being 25.8% at 40 hidden neurons. This indicates that using the first 40 SSx can reduce the amount of collected data with an acceptable reduction in the classification rate. This small reduction in classification can potentially be minimized through a more informed selection of SSx and analysis of the MCI over the entire cohort of victims.

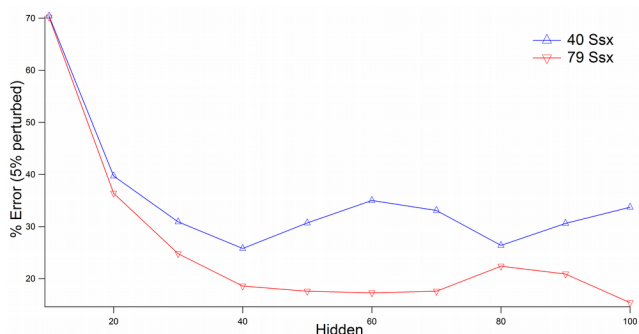

Figure 9: Optimizing the number of hidden neurons used in training the Artificial Neural Network. We used the 5% perturbed simulated patient-data for additional testing on the model.

## 4 Conclusion

Our overall approach consisted of evaluating WISER in application to MCI under more realistic conditions. We have used the results of WISER as the basis of comparison to highlight the advantages and disadvantages of BDT and ANN, two common classification approaches in machine learning. The summary of results shown in Figure 10 illustrates the significant improved chemical identification performance that can be obtained from BDT or ANN compared to WISER. Results reported in section 3.1 (also summarized in Figure 10) reflect the intolerance of WISER to erroneous and imperfect data; a condition that is very likely to occur during the chaos and confusion that occurs during an MCI. Furthermore, WISER operates with a luxury of reporting a potentially long list of unrelated chemicals that share a common list of present SSx. Presenting a long list of unrelated chemicals may provide additional confusion during an MCI. However, creating a list of chemicals affords the benefit of operating with fewer SSx. Therefore, WISER exhibits the advantage of using as many or as little number of SSx as are available while BDT and ANN require a fixed number of SSx in their successful deployment.

Results for BDT and ANN evaluations reported in sections 3.2 and 3.3 highlight the significant robustness of these more sophisticated approaches compared to WISER. In summary, BDT and ANN show promise when compared to WISER for quickly and accurately identifying a culprit chemical during a chemical MCI. This gain in robustness is achieved through the use of these machine-learning techniques' ability to generalize and not simply memorize. Furthermore, BDT provides the clear advantage of arriving at a single chemical with requiring only 9 SSx (based on the depth of the tree shown in Figure 7). ANN exhibited the same degree of robustness compared to the BDT but with the apparent disadvantage of requiring all 79 SSx during the process of substance identification. However our exploration of dimensional reduction and results shown in section 3.4 support the possibility of using only 40 of the 79 SSx with little reduction in performance.

Our future investigations will focus on further reduction of data dimensionality by the use of previously established methods such and Principal Component Analysis (PCA) or Linear Discriminant Analysis[15] (LDA). AI tools employed during chemical MCIs could dramatically reduce the amount of information collected from patients resulting in increased accuracy, precision, and efficiency in identifying the chemical.

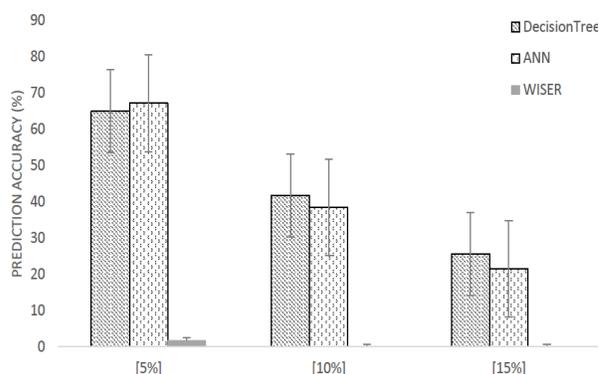

Figure 10. Overall prediction accuracy for each model tested with 31,100 simulated patient-data perturbed at 5%, 10% and 15%

## 5 Acknowledgments

This study was supported by the National Institutes of Health/The National Library of Medicine grant number 5R01LM011648.